\documentclass[10pt,twocolumn,letterpaper]{article}

\usepackage{wacv}
\usepackage{times}
\usepackage{epsfig}
\usepackage{graphicx}
\usepackage{amsmath}
\usepackage{amssymb}
\usepackage{comment}
\usepackage[accsupp]{axessibility}

\def\wacvPaperID{0003} 

\wacvfinalcopy 

\ifwacvfinal
\fi

\ifwacvfinal
\usepackage[breaklinks=true,bookmarks=false]{hyperref}
\else
\usepackage[pagebackref=true,breaklinks=true,colorlinks,bookmarks=false]{hyperref}
\fi

\begin{document}

\title{Explainability of the Implications of Supervised and Unsupervised Face Image Quality Estimations Through Activation Map Variation Analyses in Face Recognition Models}


\author{Biying Fu$^{1}$, Naser Damer$^{1,2}$
\\
$^{1}$Fraunhofer Institute for Computer Graphics Research IGD,
Darmstadt, Germany\\
$^{2}$Department of Computer Science, TU Darmstadt,
Darmstadt, Germany\\
Email: {biying.fu@igd.fraunhofer.de}
}

\maketitle
\ifwacvfinal
\thispagestyle{empty}
\fi

\begin{abstract}
It is challenging to derive explainability for unsupervised or statistical-based face image quality assessment (FIQA) methods. In this work, we propose a novel set of explainability tools to derive reasoning for different FIQA decisions and their face recognition (FR) performance implications. 
We avoid limiting the deployment of our tools to certain FIQA methods by basing our analyses on the behavior of FR models when processing samples with different FIQA decisions.
This leads to explainability tools that can be applied for any FIQA method with any CNN-based FR solution using activation mapping to exhibit the network's activation derived from the face embedding. To avoid the low discrimination between the general spatial activation mapping of low and high-quality images in FR models, we build our explainability tools in a higher derivative space by analyzing the variation of the FR activation maps of image sets with different quality decisions.
We demonstrate our tools and analyze the findings on four FIQA methods, by presenting inter and intra-FIQA method analyses. 
Our proposed tools and the analyses based on them point out, among other conclusions, that high-quality images typically cause consistent low activation on the areas outside of the central face region, while low-quality images, despite general low activation, have high variations of activation in such areas. 
Our explainability tools also extend to analyzing single images where we show that low-quality images tend to have an FR model spatial activation that strongly differs from what is expected from a high-quality image where this difference also tends to appear more in areas outside of the central face region and does correspond to issues like extreme poses and facial occlusions. The implementation of the proposed tools is accessible here \footnote{\url{https://github.com/fbiying87/Explainable_FIQA_WITH_AMVA}}.

\end{abstract}

\section{Introduction}
Face recognition (FR) systems are becoming more widely used in our daily life, be it for security-relevant areas such as border control or for unlocking your personal devices such as smartphones. This spread relates highly to the performance improvements due to advances made in deep-learning methods for FR \cite{deng2019arcface,DBLP:conf/icb/BoutrosDFKK21,DBLP:journals/corr/abs-2109-09416}. 

Low utility \cite{ISOIEC29794-1} face samples largely effects the performance of FR algorithms \cite{best2018learning}.
Therefore, to improve the performance of FR and overcome the aforementioned challenges, advances have been made to enhance the performance and dependability by choosing to use high-quality face images. 
Face image quality is also used to weight face embeddings and comparison scores when performing multi-frame (video)  \cite{meng2021magface,DBLP:conf/icpr/DamerSN14,ijbc} or multi-spectrum \cite{DBLP:conf/fusion/MallatDBD19} face verification.
Facial image quality assessment (FIQA) methods have been developed to evaluate this metric for facial images. Recent advances in the development of FIQA methods with deep-learning-based approaches already show good results in improving the FR performance. Few works focus also on building FIQA methods based on interpretable reasoning in terms of uncertainty \cite{DBLP:conf/iccv/ShiJ19}, embedding robustness \cite{terhorst2020ser}, or
embedding magnitude that correlate to the sample location with respect to its class
\cite{meng2021magface}. However, little effort was put into the explainability of these methods.

To address this unscouted field, we develop a set of novel explainability tools.  
To start with, and to enable our tools to be deployed to both supervised and unsupervised FIQA approaches, we do not look into the responses of the FIQA approach itself, but rather the FR model behavior when processing images with various FIQ decisions. 
Given that the general spatial activation of FR models to low or high-quality images is rather similar, we take this analysis to the higher derivative level by looking at the activation map variation analyses.
The newly proposed tools successfully target answering three questions related to the explainability of FIQA decisions. These questions address the following issues: 1) What makes the face images of high quality (in comparison to low) in the view of different FIQA based on the FR model behavior, 2) what makes the decision of high or low quality different between different FIQAs based on the FR model behavior, and 3) how do the activation mappings caused by face images in the FR model deviate from what is expected from high-quality images and how does that reflect in their quality score.

The structure of our work is as follows: we first introduce the related work in terms of FIQA methods in Section \ref{sec:related}. In section \ref{sec:methodology}, we introduce our proposed explainability tools. The experimental setup is described in Section \ref{sec:experimental_setup}, while the analyses based on our proposed tools are followed in Section \ref{sec:results}. We conclude our paper in Section \ref{sec:conclusion} where we shortly summarized our main findings.

\section{Related Works}
\label{sec:related}

The estimation of face images utility with FIQA methods helps FR systems to improve their performance by avoiding the processing of low-quality, and thus low-utility captures.

Most recent works of FIQA methods mainly focused on enhancing the performance of the proposed metric in terms of the FR accuracy when neglecting low-quality samples. 
An example of that is the probabilistic Face Embeddings (PFEs) proposed by Shi et al.~\cite{DBLP:conf/iccv/ShiJ19} where they represented each face image as a Gaussian distribution in the latent space, where the variance of the Gaussian indicates the uncertainty in the feature space. This uncertainty is used as a measure of quality. 
Another example is the SDD-FIQA \cite{DBLP:journals/corr/abs-2103-05977} where Ou et al.~proposed a supervised method using generated quality pseudo-labels by calculating the Wasserstein Distance between the intra-class similarity distribution and inter-class similarity distributions of sample identities. With these peseudo-labels, the network trains a regression network for the quality prediction. Other FIQA metrics using the trustworthiness as in SER-FIQ \cite{terhorst2020ser} or MagFace \cite{meng2021magface} by learning a universal feature embedding which magnitude is a direct measure of quality, are further introduced in Section \ref{sec:experimental_setup}. 

Recent works have tried to have a detailed look into the contribution of facial parts to the estimated face image quality (FIQ). 
Fu et al. \cite{fu2021relative} investigated the different face sub-regions (including eyes, mouth, and nose) and showed their relative importance towards face utility by comparing general image quality assessment (IQA) metrics on these areas. 
However, this work only looked at IQA metrics and thus does not provide many insights on the explainability of FIQA. 
Further works have looked into the spatial activation maps of supervised FIQA and IQA methods and analyzed how they are affected when facing different sample categories such as low and high-quality \cite{DBLP:journals/corr/abs-2110-11111}, or even masked faces \cite{DBLP:journals/corr/abs-2110-11283}.
However, these efforts were limited to the supervised FIQA methods and did not address the better performing unsupervised FIQA methods, as the nature of the unsupervised approaches does not allow for rational activation map analyses.
Moreover, none of the previous FIQA methods tried to provide an explaination of a quality decision by looking into the spatial interpretation of the response of FR models to what is deemed as low or high-quality samples, rather than just analyzing the FIQA behavior.  
In this work, we propose a generalized methodology to enhance the explainability of the behavior of FR models on low and high-quality samples, as well as face image quality estimation decisions, by analyzing the FR activation mappings, rather than these of FIQA.

\section{Methodology}
\label{sec:methodology}

In this section, we propose a novel set of tools to explain the face image quality and its effect on FR model behavior, independent of the underlying working principles of the FIQA methods itself, by focusing on the response of the FR models to samples labeled with different qualities. 
We leveraged the process of activation mapping of a visualization network to display the scaled activation weighted by the face embedding of the FR network. This mapping links the content of face embeddings with the pixels in the input face image. Based on the activation mapping, we draw statistic characteristics for images with different face qualities and thus enable analyses of the response of FR models to low and high face image qualities and the quality interpretation of single samples based on FR model responses.   

To illustrate our proposed method with a concrete example, let us assume that we chose MagFace~\cite{meng2021magface} as the underlying unsupervised FIQA method and used the ResNet-100 \cite{han2017deep} model trained with ArcFace loss \cite{deng2019arcface} as the face recognition model to extract the face embeddings. We further used ScoreCAM~\cite{wang2020score} as the approach for the activation mapping process. The activation mapping visualized the deepest convolution layer of the Res-Net100 and upsampled it to overlay to the input layer. The scaled version of the activation measures how the output changes to the face embedding. For each image, the activation mapping (AM) provided an output activation map with each pixel value noted as $a_{i,j},i=1:112,j=1:112$ of the size 112\text{x}112. However, this concept can be extended to any FIQA method and CNN-based FR model. More details about the exact models used in this work and the reason for this selection are provided later in Section \ref{sec:experimental_setup}.

Using the selected FIQA metric, e.g., MagFace, we calculate the face image scores for a given face images database. The calculated face image scores are used as ground truth to determine the group of low and high-quality images. In our experiment, we chose the 10\% of face images with the lowest and 10\% of face images with the highest FIQ scores. For simplicity, these two groups are named H and L individually. 

For each of these two individual groups of H and L images, we introduce and define the mean activation mappings (MAM) and denote them as MAM$_H$ and MAM$_L$. These maps have the same dimension as the input with 112\text{x}112 pixels. Each element in the MAM is noted as $\overline{a_{i,j}}$ and it is derived from equation (\ref{eq:mam}) using the activation value of each single sample $a_{i,j}$ with the running index $i=1:112$ and $j=1:112$: 
\begin{equation}
    \overline{a_{i,j}}=\frac{1}{N} \sum_{k=1}^{N} a_{i,j}^k ,
    \label{eq:mam}
\end{equation}
where $N$ is the number of images within the H or L groups respectively. As we aim to measure the variability in the activation mapping, we introduce and define the activation mapping variation map (AM-V). We further denote them as AM-V$_H$ and AM-V$_L$ respectively, for high-quality or low-quality samples. These maps also have the dimension of the input with a shape of 112\text{x}112 and each element in the AM-V is the $\text{s}_{i,j}$ and is derived according to equation (\ref{eq:am-d}): 
\begin{equation}
    \text{s}_{i,j}=\sqrt{\frac{1}{N} \sum_{k=1}^{N} (a_{i,j}^k-\overline{a_{i,j}})^2} ,
    \label{eq:am-d}
\end{equation}
where $N$ has the same meaning as in Equation (\ref{eq:mam}) and $a_{i,j}$ are extracted from elements of the activation mapping.

In order to reduce the influence of outliers in the MAM, we further looked at the median activation mapping (MDAM). The notation of MDAM$_H$ and MDAM$_L$ are also noting the MDAM for low and high quality sample sets. The element of the MDAM is denoted as $\widetilde{a_{i,j}}$ and is derived as: 
\begin{equation}
    \widetilde{a_{i,j}}=Median(a_{i,j}^k), k=1..N \text{.}
    \label{eq:mdam}
\end{equation}
We further introduce the activation mapping Median variation (AM-MV), called AM-MV$_H$ and AM-MV$_L$. The associated equation for each element of these maps $\widetilde{s_{i,j}}$ is given as: 
\begin{equation}
    \widetilde{s_{i,j}}=\sqrt{\frac{1}{N} \sum_{k=1}^{N} (a_{i,j}^k-\widetilde{a_{i,j}})^2},
    \label{eq:ammd}
\end{equation}
where both AM-MV maps have the same dimension of 112\text{x}112 pixels.

Both the defined AM-V and AM-MV maps present a visualization tool to look into the spatial areas where a relatively large variation in the activation of the FR occurs, with respect to a set of high or low-quality images. This will help identify the spatial regions responsible for the certain quality decision, despite the low consistency of these areas' location across different images.

As will be shown later, the differences between the MAM (or MDAM) of image sets of different qualities do not uncover a lot of explainability information. Therefore, to uncover the spatial related differences between these groups, we rather analyze the differences between the variations in the activation mapping (AM-V or AM-MV). We introduce these differences as the Differential activation mapping variation (D-AM-V) as in (\ref{eq:D-AM-V}) and the Differential activation mapping Median variation (D-AM-MV), as in  (\ref{eq:D-AM-MV}):
\begin{equation}
    \text{D-AM-V} = |\text{AM-V}_H - \text{AM-V}_L| ,
    \label{eq:D-AM-V}
\end{equation}
and
\begin{equation}
    \text{D-AM-MV} = |\text{AM-MV}_H-\text{AM-MV}_L|,
    \label{eq:D-AM-MV}
\end{equation}
where both equations (\ref{eq:D-AM-V}) and (\ref{eq:D-AM-MV}) can be extended to look at the differences of variations of any sets of images, not only L and H, but also to sets of images determined to be H or L by different FIQA approaches.

The proposed visualization maps provide a useful tool to analyze the differences in FR model responses to sets of facial images belonging to different sets, here sets with different FIQ determined by any FIQA approaches, or sets of a certain FIQ label determined by different FIQA approaches.

So far we introduced methods to visualize and analyze the behavior of FR models between sets of images to enable a better understanding of FR response to images of different quality levels and thus understanding the used FIQA. To further analyze the quality decision of a single face image using its FR model response, we introduce the activation deviation from the MAM (AD-MAM). The elements of the AD-MAM is noted as $d_i,j$ and is calculated of an image x with its activation mapping element $a_{i,j}$ and its absolute deviation from the mean activation mapping of the high quality sets $\overline{a_{i,j}^H}$,
\begin{equation}
    d_{i,j} = |a_{i,j} - \overline{a_{i,j}^H}| ,
\end{equation}
this can be calculated for different sets of images that build the MAM, here we focus our analyses on the MAM$_H$, and thus we note our AD-MAM as AD-MAM$_H$.

\section{Experimental Setup}
\label{sec:experimental_setup}

In this section, we first introduce the face image database used to evaluate the methodology proposed in Section \ref{sec:methodology}. This was followed by the description of the three recent deep-learning-based FIQA metrics and one general IQA metric used as examples in our work. Finally, we used Res-Net100 trained with ArcFace loss as the basic FR model used as the backbone of our explainability efforts. A short experiment overview is provided before introducing the final results and more detailed analysis. 

\subsection{Database}
VGGFace2 \cite{Cao18} dataset is a large-scale database containing face images with a large variety in quality distribution which makes it a challenging FR database. The images have high diversity in poses and complex acquisition conditions. For the main analysis in this paper, we only used the official test dataset containing 500 subjects. To reduce heavy computation and a more balanced database, we randomly selected 30 face images of each subject representing the full database. This made a total of 15000 images, the list of randomly selected images from each identity will be made publicly available to enable reproducibility.

Data preprocessing includes face detection, cropping, and alignment. The Multi-task Cascaded Convolutional Networks (MTCNN)~\cite{DBLP:journals/spl/ZhangZLQ16} framework is used to detect faces from the VGGFace2. The detected faces are further cropped and aligned using similarity transform to 112\text{x}112 pixels, such that all the face images are standardized for comparison. 

\subsection{FIQA methods}

In our experiment, we selected four different quality metrics, one is an supervised general image quality called BRISQUE \cite{mittal2012no} as a baseline and three specifically designed methods for face image quality assessment, FaceQnetV1 \cite{hernandez2019faceqnet}, MagFace \cite{meng2021magface}, and SER-FIQ \cite{terhorst2020ser}, the first is a supervised FIQA and the later two are unsupervised. the following introduces these methods shortly.

\textbf{BRISQUE} \cite{mittal2012no} proposed by Anish et al.~in 2012 is an opinion-aware image quality assessment method trained with human opinion scores. The assessment of the image is purely based on natural scene statistics learned from distortion-generic images. The method is built on the finding by Rudermann \cite{ruderman1994statistics} that natural scene images have a luminance distribution similar to a normal Gaussian distribution. Handcrafted features were derived to quantify the deviation from the Gaussian due to image distortions. The quality estimations by BRISQUE were previously found to have a strong correlation to face image utility \cite{DBLP:journals/corr/abs-2110-11111}. 

\textbf{FaceQnet} \cite{hernandez2019faceqnet} proposed by Hernandez-Ortega et al.~in 2019 is a supervised FIQA method. The BioLab-ICAO framework was used to label the ground-truth score for the training image according to the ICAO compliance level \cite{ICAO18}. This score is used to train the regression layer on top of the feature extraction layers. FaceQnet is based on fine-tuning a pre-trained FR network (RseNet-50) and the successive regression layer to associate an input image to a utility score that determines the appropriateness of the input image to an FR model. In this work, we used the latest version published in \cite{hernandez2020biometric}, i.e. FaceQnetV$_1$\footnote{https://github.com/uam-biometrics/FaceQnet}.

\textbf{SER-FIQ} \cite{terhorst2020ser} is an unsupervised deep-learning-based FIQA approach that applies stochastic variations on face representations learned from a deep-learning-based FR model by using dropout. This method mitigates the need for any automated or human labeling. The face image was passed to several sub-networks of a modified FR network by using different dropout patterns. Images with high utility are expected to possess similar face representations resulting in low variance. Thus, this proposed metric linked the robustness of face embeddings directly with FIQ. 

\textbf{MagFace} \cite{meng2021magface} by Meng et al. is another recently proposed unsupervised FIQA method based on using the adaptive loss incorporating the face image quality to the magnitude of the face embedding. This method can derive both the face representation and the face image quality from calculating the magnitude of the face embedding. The loss optimizes the inter-class variability and intra-class similarity. The MagFace version used in this work is trained on MS-Celeb-1M database and used the ResNet-100 as the backbone. 

\begin{figure}
    \centering
    \includegraphics[width=.45\textwidth]{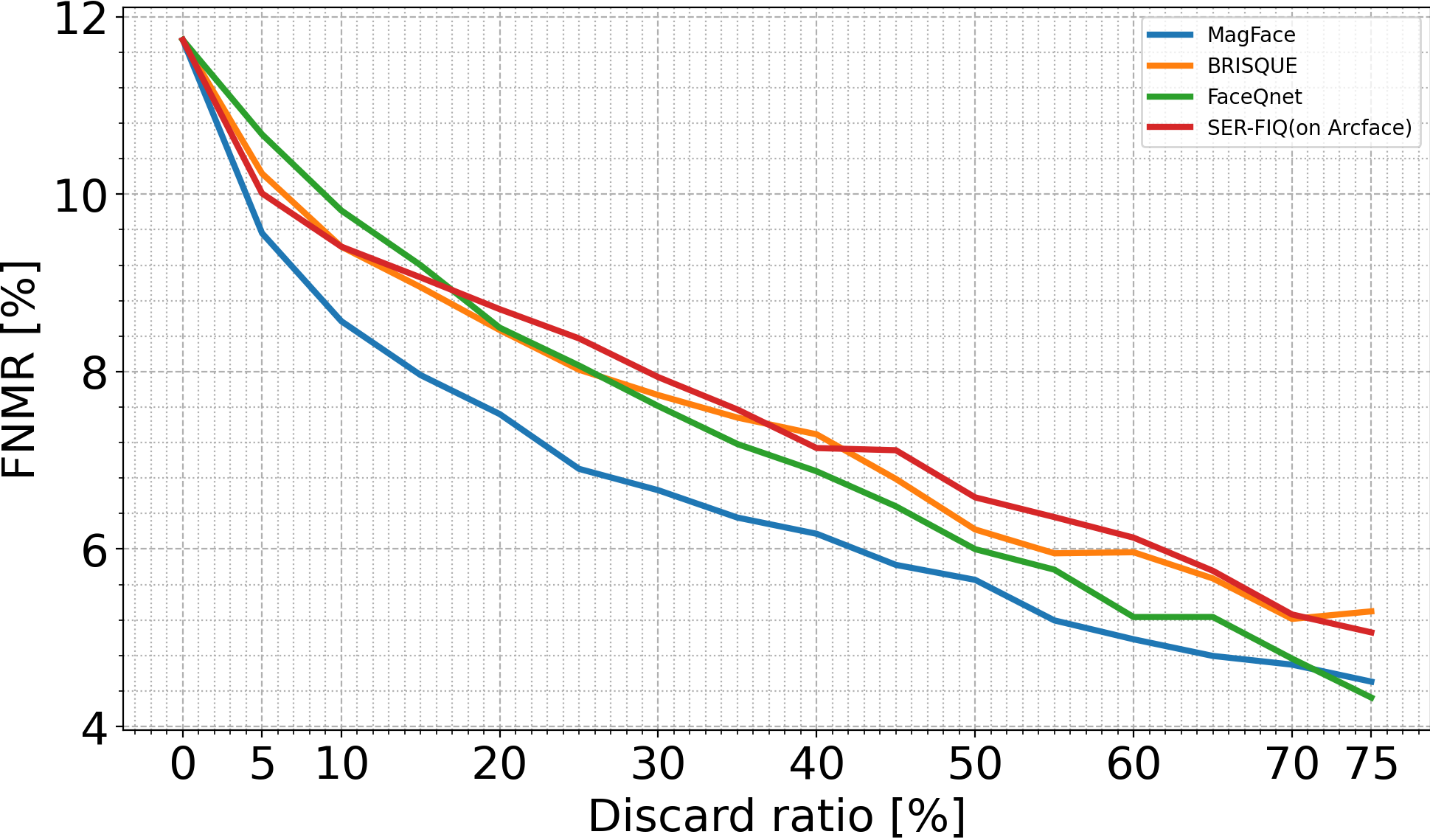}
    \caption{Error versus reject curve shows the FNMR at fixed FMR=0.1\% decays with increasing number of worse quality images are discarded. All four FIQA methods seem to perform well as a strong decrease in error is observed when the predicted low-quality images were removed.}
    \label{fig:erc_001}
\end{figure}

All selected FIQA methods perform well on the VGGFace2 database. This can be seen in the error versus reject characteristic (ERC) presented in Figure \ref{fig:erc_001}. The ERC shows the relative performance of the FR system when rejecting different ratios of the evaluation data with the lowest quality according to each FIQA metric. Figure \ref{fig:erc_001} shows the ERC with the false non-match rate (FNMR) at different ratios of rejected (low quality) images using a fixed false-match rate (FMR) at 0.1\%. The error clearly decreased as the number of worst quality samples are discarded. 

\begin{figure}
    \centering
    \includegraphics[width=.45\textwidth]{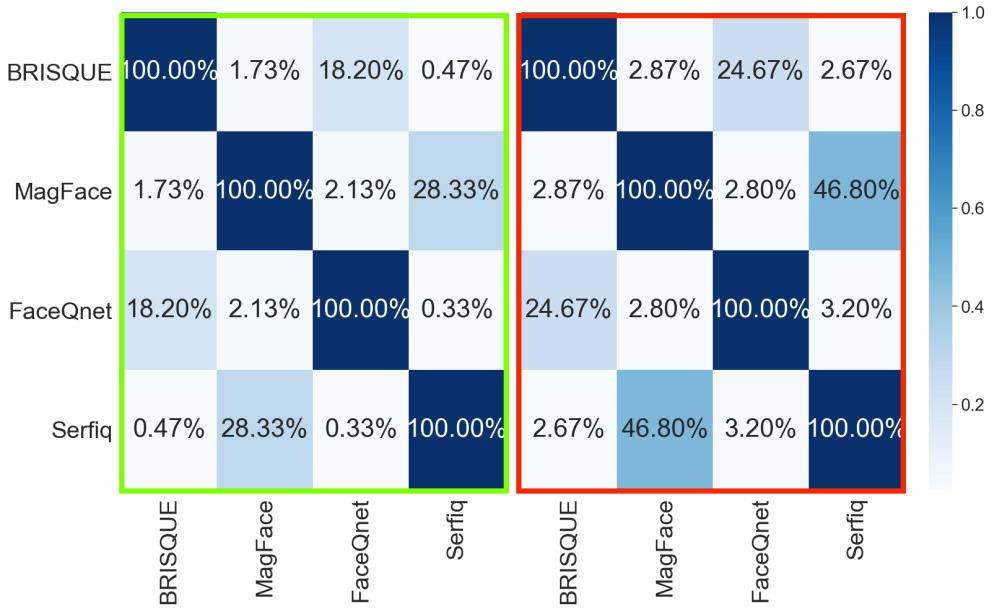}
    \caption{We displayed the samples' overlap ratio between the samples of the highest quality (10\% of the data) on the right between every pair of quality estimation methods, and the same for the 10\% of the lowest quality on the left.}
    \label{fig:confusion_matrix}
\end{figure}

In the overlapping ratio matrix shown in Figure \ref{fig:confusion_matrix}, we see further that the set of the lowest and highest 10\% sample images are not fully identical for different FIQA methods, indicating that the proposed metrics in Section \ref{sec:methodology} are derived from different base samples. The largest overlap is found for both unsupervised FIQA methods, i.e. for MagFace and SER-FIQ.

\subsection{Face Recognition Solution}
We use the ResNet-100 trained with ArcFace loss as the main FR solution to visualize the activation mapping of the input face images. This \textbf{ArcFace} \cite{deng2019arcface} model is trained using the MS1M dataset \cite{guo2016ms}. The loss function applied additive angular margin to improve the discriminative power of the FR model. We chose this model because of its improved accuracy on LFW \cite{huang:inria-00321923} 99.83\% and YTF DB \cite{wolf2011face} 99.02\%.

\subsection{Activation mapping method}
ScoreCAM \cite{wang2020score} is used as the activation mapping method to display the activations of the deepest convolution layer of the Res-Net100 and upsampled to overlay to the input layer. This choice is motivated by the extensive successful use of ScoreCam as an activation mapping tool in various biometric domains \cite{FANG2022108398,DBLP:conf/icb/FangDBKK21}. ScoreCAM provides a scaled version of the activation for the face embeddings. The weights of the scale factors are derived from this embedding. We chose ScoreCAM as it seems to provide more realistic explainability analyses to other methods such as CAM \cite{DBLP:conf/cvpr/ZhouKLOT16} and Grad-CAM \cite{DBLP:conf/iccv/SelvarajuCDVPB17}.

\subsection{Experiment overview}
Given the methodology proposed in Section \ref{sec:methodology} and the experimental setup in Section \ref{sec:experimental_setup}, we intend to show: (1)  difference between high and low-quality decision of FIQA based on AM-V/AM-MV and D-AM-MV, (2) differences within the low and high-quality decisions across FIQA methods, and (3) individual sample quality explainability with its AD-MAM of the individual FIQA method.

All four FIQA metrics were used to determine the face image quality of the used database. Out of which we determine the 10\% of the face images with the highest and lowest face qualities to the group H and L, which makes 1500 face images in each of these two groups. 
ScoreCAM builds the activation mapping from the FR using ArcFace model for these input images of both groups H and L. For the FR model trained with ArcFace loss, we used the official Pytorch version from the official Github \footnote{https://github.com/deepinsight/insightface}. 

\section{Results and Analyses}
\label{sec:results}

This section is structured in three main parts related to explaining FIQ estimation within and across the decisions of different FIQA methods. The methodologies used as explainability tools are introduced in Section \ref{sec:methodology}. 

\paragraph{1. What makes the face images of high quality (in comparison to low) in the view of different FIQA based on the FR model behaviour?}

\begin{figure*}
    \centering
    \includegraphics[width=.95\linewidth]{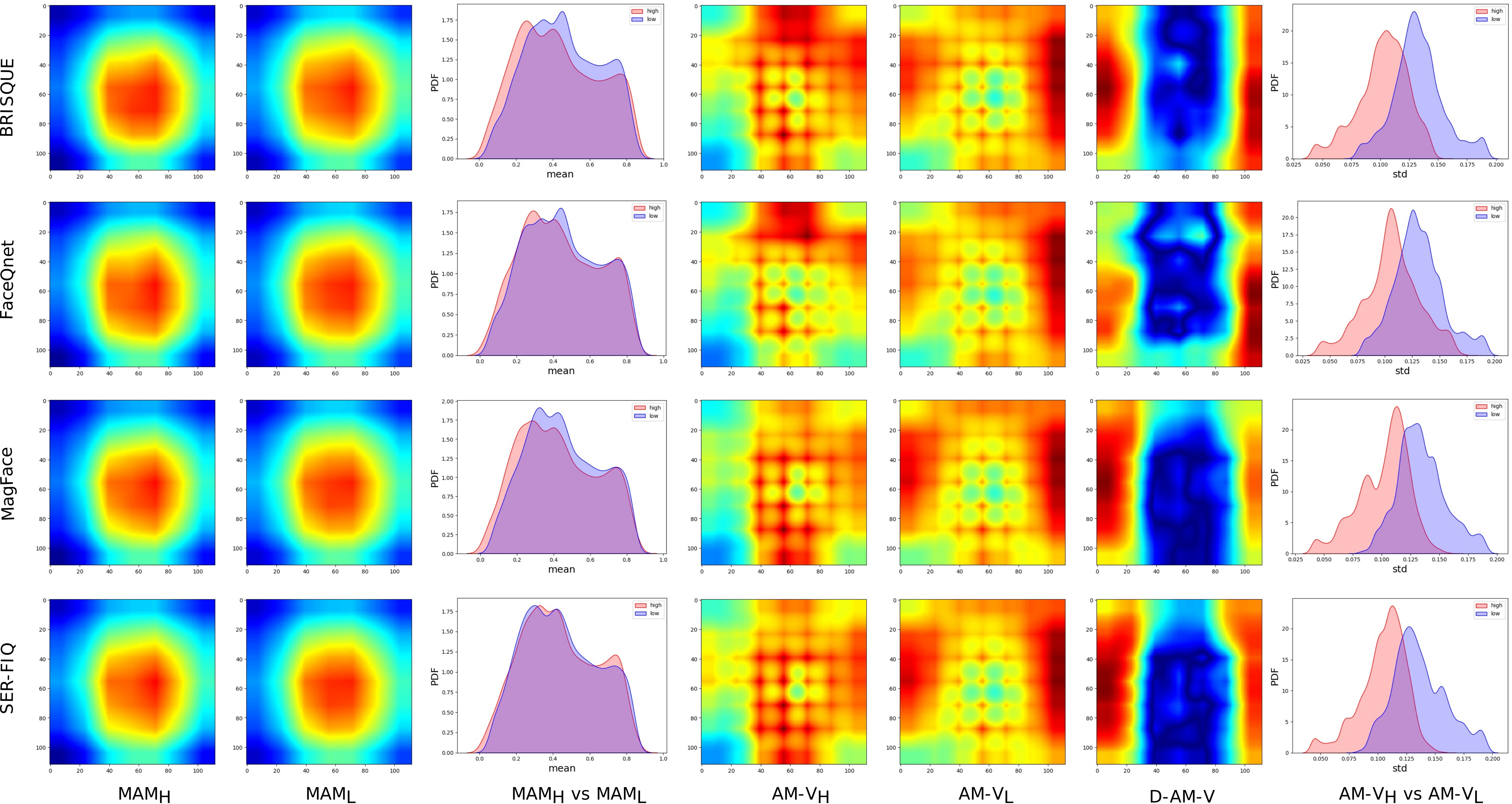}
    \caption{The results are derived from the MAM for each FIQA metric individually. Even though the MAM$_H$ and MAM$_L$ are similar for both H and L groups, a strong deviation can be observed in terms of AM-V, indicating low-quality images have a stronger deviation from the mean compared to high-quality images, especially on image borders. It is to be noted that for the visualization purpose, these deviation maps AM-V$_H$, AM-V$_L$, and D-AM-V are scaled.}
    \label{fig:intra_method_statistic_mean}
\end{figure*}
\begin{figure*}
    \centering
    \includegraphics[width=.95\linewidth]{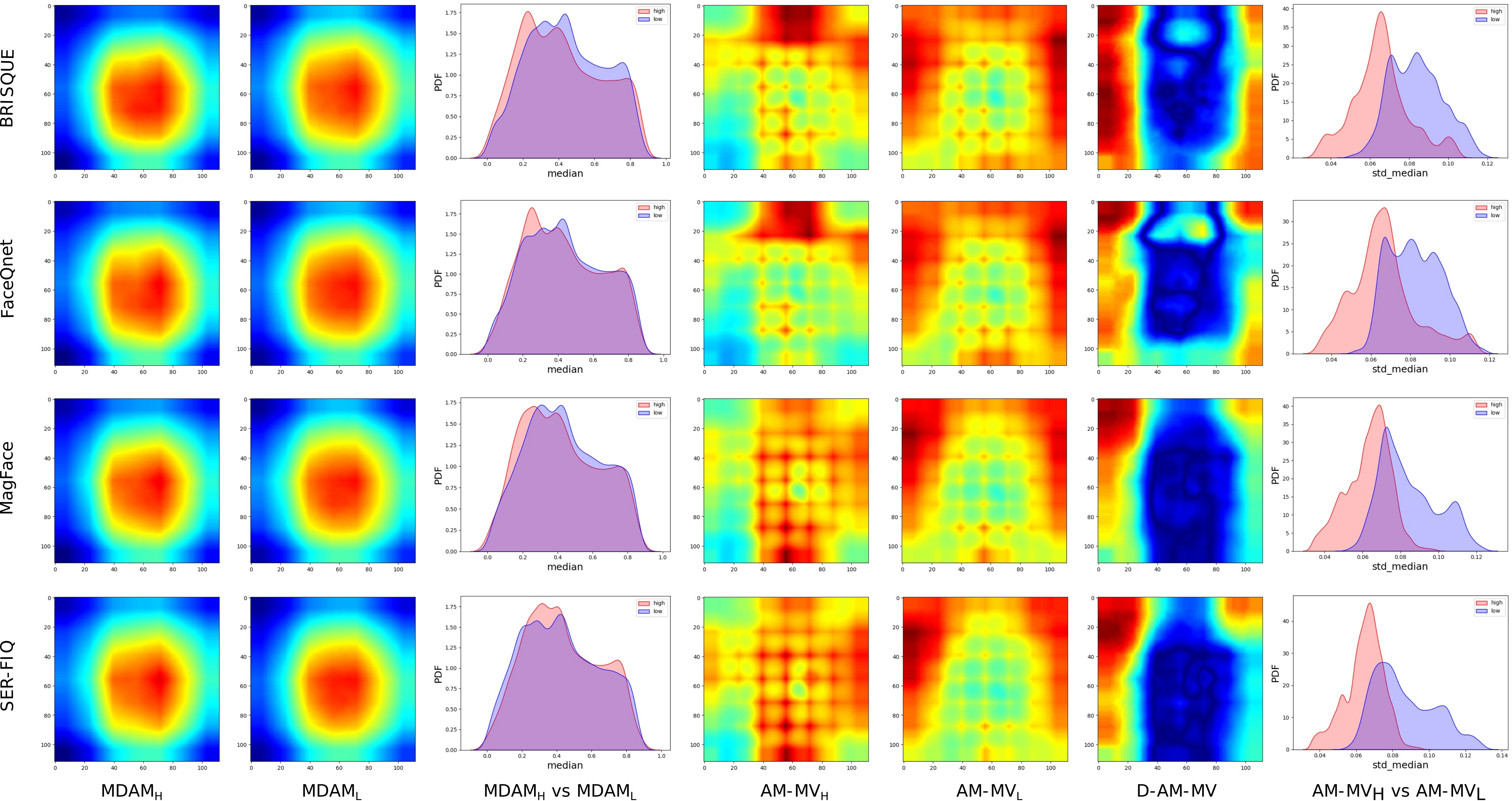}
    \caption{The results are derived from the MDAM for each FIQA metric individually. The same result is obtained as for MAM in Figure\ref{fig:intra_method_statistic_mean}, where even stronger variations are observed from the median for low-quality images. Low-quality images tend to attract the attention of FR model in areas outside of the center
face area, which can be caused by different reasons like e.g., postures or occlusions. The variation maps are re-scaled for better visualization.}
    \label{fig:intra_method_statistic_median}
\end{figure*}

Figure \ref{fig:intra_method_statistic_mean} and Figure \ref{fig:intra_method_statistic_median} depict the results for the MAM and MDAM for each FIQA method individually,  we do not notice major differences between MAM and MDAM. Taking a look at the distributions for the mean activation mapping for H and L, no visible significant differences are noticed. This acts as the major motivation behind our proposed analyses based on activation map variation, rather than the activation maps themselves. Even though the difference for the MAM/MDAM between the H and L sets is not strongly visible, there are strong variations in the AM-V/AM-MV. Comparing the value distributions for AM-V$_H$ and AM-V$_L$, we clearly observe that the variations for L are significantly larger compared to H, the same can be observed for AM-MV$_H$ and AM-MV$_L$. This might indicate that the variability in activation mapping is stronger for low-quality face images. 
Looking at the AM-V/AM-MV for L and H, we noticed that the values are typically higher in L on the borders of the images, while it is higher in the middle face region for H. This indicates that the lower utility of L is based on the FR model focusing more often on the border areas rather than the center, in comparison to the H set.
This result is further confirmed by looking at the D-AM-V mapping, where we can see that stronger deviations are observed on the left and right borders of the face image.

Generally, all considered FIQA methods lead to similar MA-V/AM-MV and D-AM-V/D-AM-MV observations, indicating that the effect of quality differences (as per different FIQAs) on the FR model is of the same nature.
Based on these observations and to answer the question driving this subsection, we can notice that despite the similarity of the general activation maps of low and high-quality images (similar MAM/MDAM for L and H), what makes an image high quality is the consistent low activation on the areas outside of the face center, while low-quality images, despite general low activation in these areas, have high variations of activation there. In simple words, low-quality images do attract the attention of FR model in areas outside of the center face area, however, in different locations and less consistently, which can be caused by the different reasons for the degradation of quality.

\paragraph{2. What makes the decision of high or low quality different between different FIQAs based on the FR model behaviour? }

\begin{figure*}
    \centering
    \includegraphics[angle=90, width=.95\linewidth]{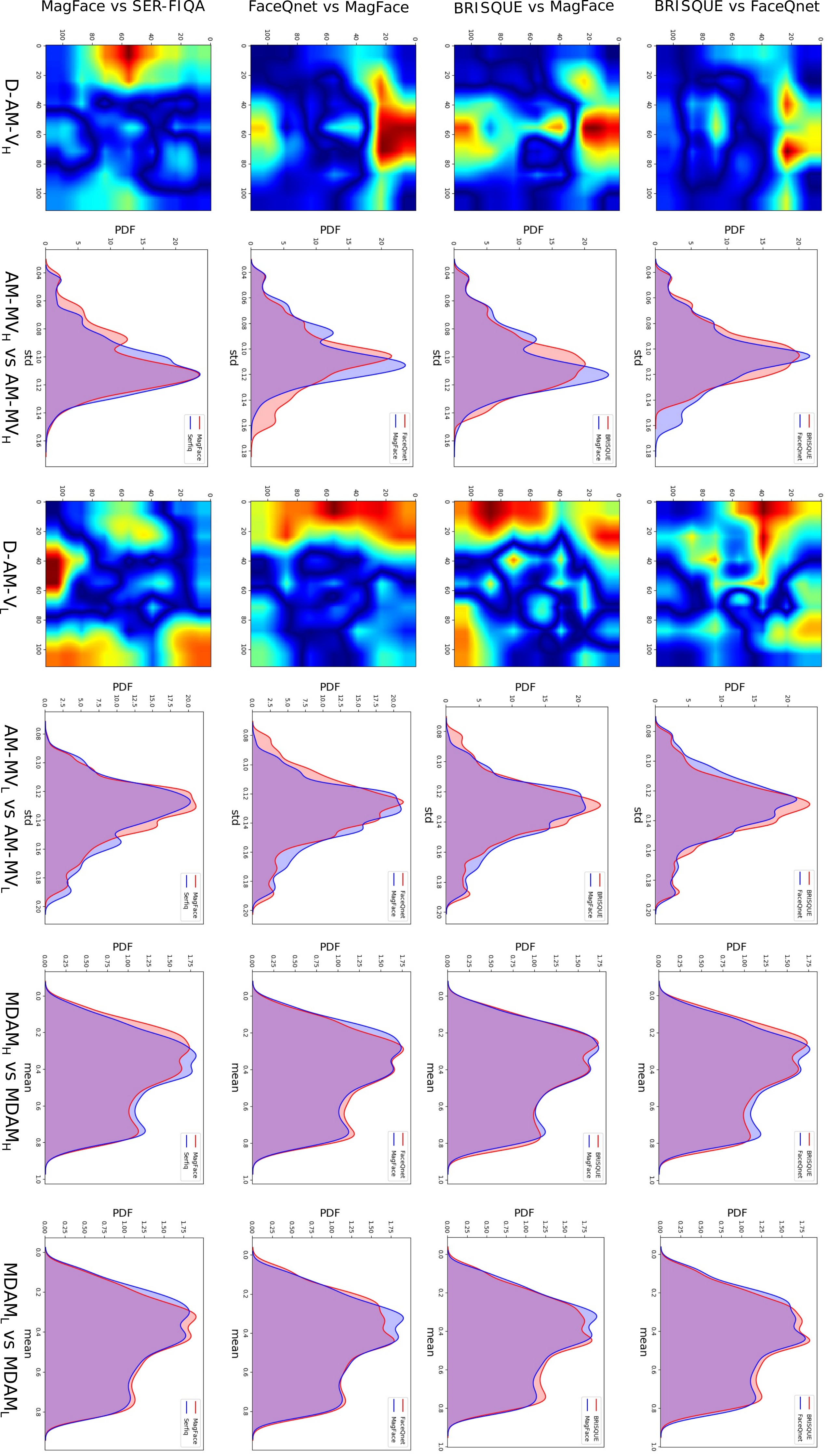}
    \caption{Results are derived when comparing D-AM-V for H and L sets across different FIQA methods. D-AM-V$_H$ and D-AM-V$_L$ between the unsupervised MagFace and the supervised methods FaceQnet and BRISQUE show similar tendencies. D-AM-V$_H$ and D-AM-V$_L$ tend to show stronger and more consistent differences between supervised and unsupervised methods, hinting a link between the lower performance of FaceQnet and BRISQUE and the focus of the FR model on areas outside of the face center in an unexpected manner .}
    \label{fig:cross_method_statistic}
\end{figure*}

In Figure \ref{fig:cross_method_statistic} we see visualization of our D-AM-V mapping across FIQA methods. The D-AM-V here differs than the one presented in Equation (\ref{eq:D-AM-V}) by looking across FIQA methods rather than quality sets, here the 
$\text{D-AM-V}_H = \text{AM-V}_{H,FIQA-1} - \text{AM-V}_{H,FIQA-2}$ and 
$\text{D-AM-V}_L = \text{AM-V}_{L,FIQA-1} - \text{AM-V}_{L,FIQA-2}$.
D-AM-V$_H$ and D-AM-V$_L$ show the differential activation mapping variations for H and L sets individually between pairs of FIQA metrics. 
 D-AM-V$_H$ and  D-AM-V$_L$ between the unsupervised MagFace and the supervised methods FaceQnet and BRISQUE show similar tendencies. Samples selected to be H by the supervised methods tend to cause higher activation variations in the top and bottom of the image when compared to MagFace. Samples selected to be L by the supervised methods tend to cause higher activation variations on the left and right edges of the face image when compared to MagFace. This consistent activation variation on areas out of the face center can rationalize the performance differences between FaceQnet and BRISQUE on one side, and MagFace on the other side, see Figure \ref{fig:erc_001}. There are fewer differences in the activation variations caused by both the H and L set between the supervised (and with poorer performance) FIQA methods (BRISQUE and FaceQnet) both in terms of magnitude (distribution shifts) and clear spatial distribution. The same can be seen between the two high-performing unsupervised methods (SER-FIQ and MagFace). This leads to answering the question behind this sub-section by stating that compared to high performing FIQA methods, lower performing FIQA methods have larger FR activation variation on the edges of the face image. 


\paragraph{How do the activation mappings caused by face images in the FR model deviate from what is expected from high quality images and how does that reflect in their quality score?}

\begin{figure*}
    \centering
    \includegraphics[angle=90, width=.98\linewidth]{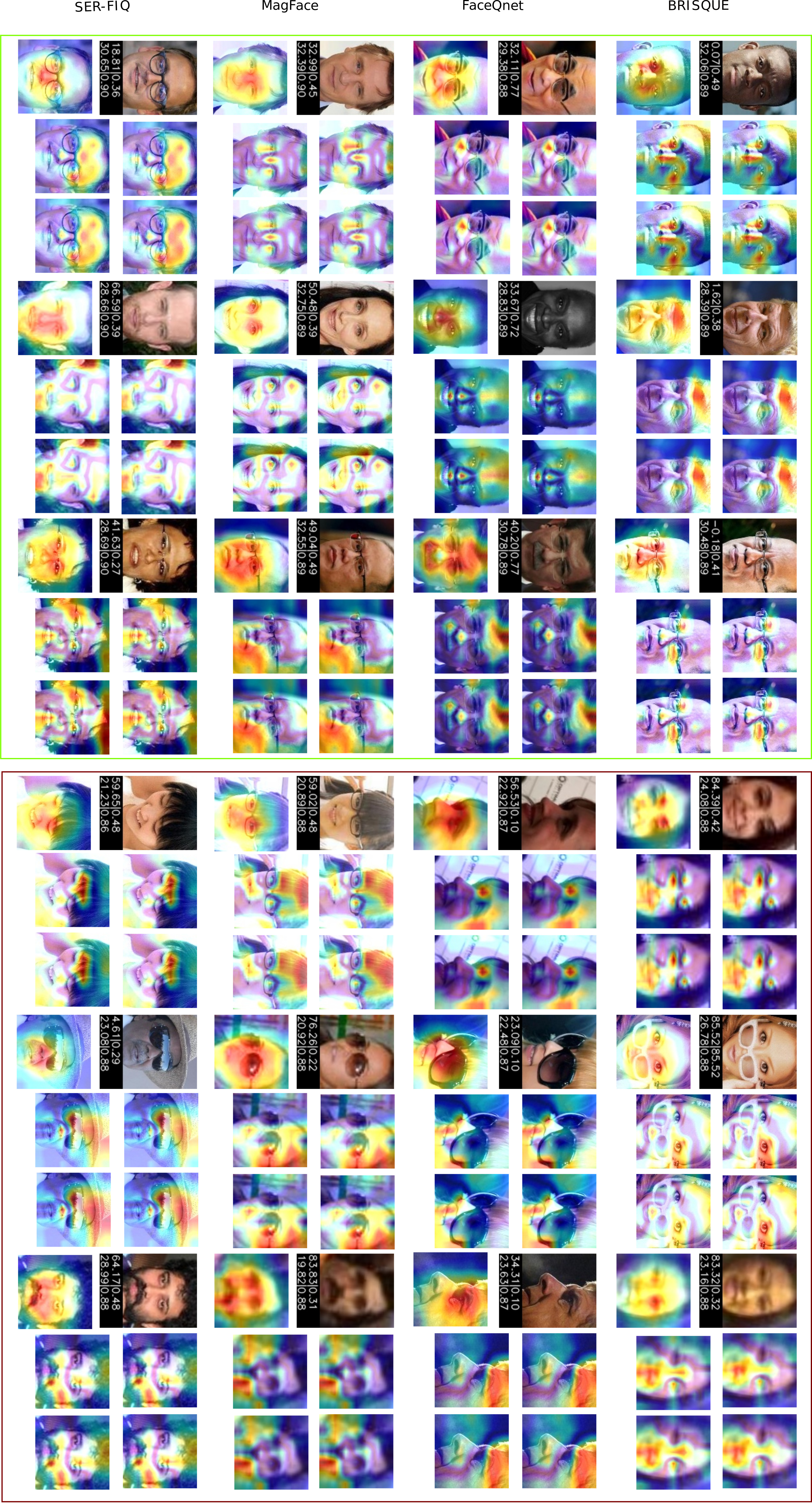}
    \caption{Selected face images with high and low face qualities are displayed. Each sample subject contains one original image with quality scores illustrated below, one image overlapped with the activation mapping derived from ArcFace FR, and four overlayed AD-MAM$_H$ maps. These differential maps indicate the deviations from the MAM$_H$ and show areas which could cause the degradation in qualities, such as sunglasses, hat, and mustaches.}
    \label{fig:samples}
\end{figure*}

In Figure \ref{fig:samples} we depicted sample images of high and low face qualities. Each sample subject, we provided one original image with the FIQ score from all four FIQA metrics below, one image overlayed with the activation mapping from the ArcFace FR solution, and four overlayed with the AD-MAM$_H$ map of each FIQA method. These differential activation mappings are for BRISQUE, FaceQnet, MagFace, and SER-FIQA in the correct ordering starting from upper left to bottom right displayed for each sample subject. These AD-MAM$_H$ maps show areas which could cause the degradation in qualities as they deviate from the mean template and is emphasized in the visualization.

From Figure \ref{fig:samples}, it can be generally concluded that the AD-MAM$_H$ for all FIQA methods contain larger and higher values for low quality images in comparison to high quality images. Also in this comparison, the high value areas in the AD-MAM$_H$ for low quality images tend to appear more in the areas around the face rather than the center of the face. This is less apparent in the high quality images. 
This larger and higher values in AD-MAM$_H$ corresponds to the low quality estimated across the four FIQA methods and in many cases it is related to less than optimal poses, face occlusions, and overall low image sharpness.
To answer the question motivating this sub-section, our analyses and proposed explainability tools reveal that low quality images tend to have a FR model activation map that strongly differs than that of what is expected from a high quality image. This difference also tends to appear more in the areas outside of the central face region. 



\section{Conclusion}
\label{sec:conclusion}

Making the face image quality estimation explainable is a challenging task that goes beyond analyzing the FIQA network itself. Most recent works put more focus on designing FIQA methods that perform well without looking into what is the response of an FR model to high or low-quality face image. In this work, we presented a novel set of explainability tools to enhance the visual explainability of FIQ estimation decisions based on the variation analyses in FR models. The proposed tools can be applied for any FIQA method with any CNN-based FR solution using activation mapping to exhibit the network's activation derived from the face embedding. 
By showing the intra-groups and cross-method inter-groups statistics of the network's activation, we try to relate explainability to groups of H and L face quality image sets for the individual FIQA method. We demonstrate that even though the MAM between H and L is small, the variations in activation mapping for L are larger compared to H.
This points out that the low-quality images tend to cause the FR network to focus on areas outside of the central face area, however, in an inconsistent manner, as the reason causing the low quality can vary largely.
We additionally link this observation to the relative performance of different supervised and unsupervised FIQA approaches. Finally, we look at the explainability of the quality decision of individual face images by analyzing the differences between their activation maps in FR models and the maps expected from high-quality images, pointing out consistent differences in FR behavior between high quality and low-quality face images.

\textbf{Acknowledgements:}
This research work has been funded by the German Federal Ministry of Education and Research and the Hessian Ministry of Higher Education, Research, Science and the Arts within their joint support of the National Research Center for Applied Cybersecurity ATHENE.




{\small
\bibliographystyle{ieee_fullname}
\bibliography{egbib}
}

\end{document}